\documentclass[sigconf]{acmart}
\usepackage{times}
\usepackage{latexsym}
\usepackage{booktabs}
\usepackage[T1]{fontenc}
\usepackage[utf8]{inputenc}
\usepackage{xspace}
\usepackage{microtype}

\usepackage{algorithm}% http://ctan.org/pkg/algorithms
\usepackage{algpseudocode}% http://ctan.org/pkg/algorithmicx
\usepackage{graphicx}
\usepackage{amsmath}
\usepackage{bm}

\AtBeginDocument{%
  \providecommand\BibTeX{{%
    \normalfont B\kern-0.5em{\scshape i\kern-0.25em b}\kern-0.8em\TeX}}}

\setcopyright{acmlicensed}
\copyrightyear{2018}
\acmYear{2018}
\acmDOI{XXXXXXX.XXXXXXX}

\acmConference[Conference acronym 'XX]{Make sure to enter the correct
  conference title from your rights confirmation emai}{June 03--05,
  2018}{Woodstock, NY}
\acmISBN{978-1-4503-XXXX-X/18/06}

\acmSubmissionID{598}

\begin{document}

\title[SpeCrawler: Generating OpenAPI Specifications from API Documentation]{SpeCrawler: Generating OpenAPI Specifications from API Documentation Using Large Language Models}

\author{Koren Lazar}
\authornote{Both authors contributed equally to this research.}
\email{koren.lazar@ibm.com}
\affiliation{%
  \institution{IBM Research}
  \city{Tel Aviv}
  \country{Israel}
}

\author{Matan Vetzler}
\authornotemark[1]
\email{matan.vetzler@ibm.com}
\affiliation{%
  \institution{IBM Research}
  \city{Tel Aviv}
  \country{Israel}
}

\author{Guy Uziel}
\email{guy.uziel1@ibm.com}
\affiliation{%
  \institution{IBM Research} 
  \city{Tel Aviv} 
  \country{Israel}
}

\author{David Boaz}
\email{davidbo@il.ibm.com
}
\affiliation{%
  \institution{IBM Research}
  \city{Haifa}
  \country{Israel}
}

\author{Esther Goldbraich}
\email{esthergold@il.ibm.com}
\affiliation{%
  \institution{IBM Research}
  \city{Haifa}
  \country{Israel}
}

\author{David Amid}
\email{davida@il.ibm.com}
\affiliation{%
  \institution{IBM Research}
  \city{Haifa}
  \country{Israel}
}

\author{Ateret Anaby-Tavor}
\email{atereta@il.ibm.com}
\affiliation{%
  \institution{IBM Research}
  \city{Haifa}
  \country{Israel}
}

\renewcommand{\shortauthors}{Lazar and Vetzler, et al.}

\begin{abstract}
In the digital era, the widespread use of APIs is evident. However, scalable utilization of APIs poses a challenge due to structure divergence observed in online API documentation. This underscores the need for automatic tools to facilitate API consumption. A viable approach involves the conversion of documentation into an API Specification format.
% Previous attempts
While previous attempts have been made using rule-based methods, these approaches encountered difficulties in generalizing across diverse documentation. 
% Introduce Specrawler
In this paper we introduce SpeCrawler, a comprehensive system that utilizes large language models (LLMs) to generate OpenAPI Specifications from diverse API documentation through a carefully crafted pipeline. By creating a standardized format for numerous APIs, SpeCrawler aids in streamlining integration processes within API orchestrating systems and facilitating the incorporation of tools into LLMs. 
%by automatically generating required metadata in a concise standardized structure. 
The paper explores SpeCrawler's methodology, supported by empirical evidence and case studies, demonstrating its efficacy through LLM capabilities.
\end{abstract}






\maketitle

\section{Introduction}
\begin{figure*}
    \centering
     \includegraphics[width=0.68\linewidth]{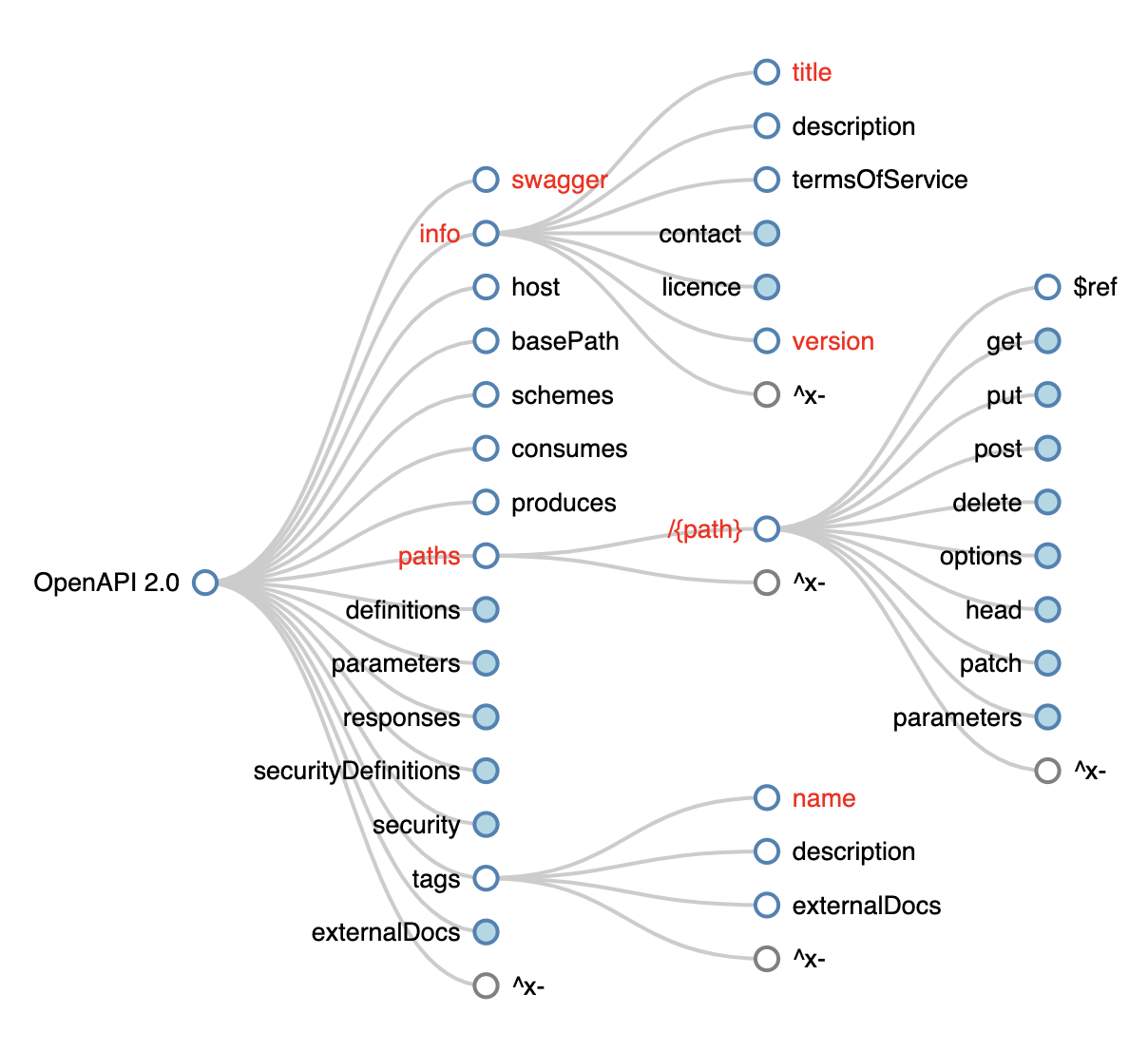}
    \caption{\label{fig:oas-schema} Illustration of an OpenAPI Specification schema structure, highlighting the hierarchical arrangement of components like paths, parameters, responses, and other elements that define RESTful APIs. Typically, OpenAPI Specification files are stored in YAML or JSON formats.
    }
\end{figure*}

% Describe what is an API Specification and what is it good for?
% Describe (shortly) OpenAPI Specification
In today's digital realm, APIs are crucial in facilitating smooth communication among varied software applications, fostering integration and interoperability. With the increasing number of APIs and their resources, standardizing their usage becomes paramount.

The OpenAPI Specification (OAS)~\footnote{\url{https://swagger.io/specification/} as accessed on [07.02.2024]} functions as a standardized structure for outlining RESTful APIs. Typically rendered in YAML or JSON formats, it meticulously outlines endpoints, request configurations, and authentication mechanisms in a machine-readable way. This helps developers to understand and interact with APIs more effectively. By establishing a common language for API description, OAS promotes interoperability, facilitates automated testing and validation of API implementations, and encourages collaboration among developers, ensuring consistency in conveying API functionalities and protocols. OAS has gained significance in guaranteeing precision and uniformity in the documentation of RESTful APIs, making it widely adopted by the industry. Incorporating the OAS framework boosts effectiveness, reduces mistakes, and cultivates uniformity across API development and integration phases. You can observe a representation of the OAS schema structure in Figure~\ref{fig:oas-schema}.

Despite its significance, crafting OASs remains a manual and labor-intensive effort, prompting an investigation into more streamlined and automated alternatives.
% Various approaches have been proposed aiming to automatically parse API documentations for their diverse and heterogeneous structure~\citep{?}. Unfortunately, these approaches have faced challenges in delivering a robust and accurate solution due to the diverse nature of such API documentation.

Various approaches have been proposed aiming to automatically parse API documentation to create OASs through rule-base approaches~\citep{cao2017automated, sohan2015spyrest}, some also integrating classic machine learning techniques~\citep{yang2018automatically, bahrami2020automated, huang2024generating}. Unfortunately, these approaches have faced challenges in delivering a robust and accurate solution due to the diverse nature of such API documentation.

% One significant challenge lies in the often complex and hard-to-parse nature of information found on API documentation websites. These sites exhibit variations in structure and the spread of information, sometimes lacking essential components. Hence, rule-based approaches which tried to navigate the intricate structures of these documents in order to parse the different elements for the API Specification, struggled to accurately address a wide range of documentation websites~\citep{bahrami2019watapi}.
Acknowledging these challenges, we present SpeCrawler, an innovative multi-stage methodology that combines rule-based algorithms and generative LLMs to automate the creation of OAS from raw HTML content of API documentation webpages. First, SpeCrawler applies rule-based algorithms informed by prior knowledge of the high-level structure of online API documentation. This step involves scraping and filtering out irrelevant information from the documentation. The filtering also ensures that we do not exceed the context length of the LLM. Then, it employs generative LLMs capable of generalizing across diverse documentation structures, thereby alleviating many assumptions about the input structure made by previous methods. To simplify the creation of the potentially intricate and nested OAS, we divide the generation process into multiple parts.

This systematic approach ensures a fine understanding of RESTful APIs' specialized languages, terms, symbols, and contexts. It enhances the system's capacity to precisely encapsulate and transform the complications of REST API documentation into comprehensive OAS. 

\begin{figure*}
    \centering
     \includegraphics[width=0.9\linewidth]{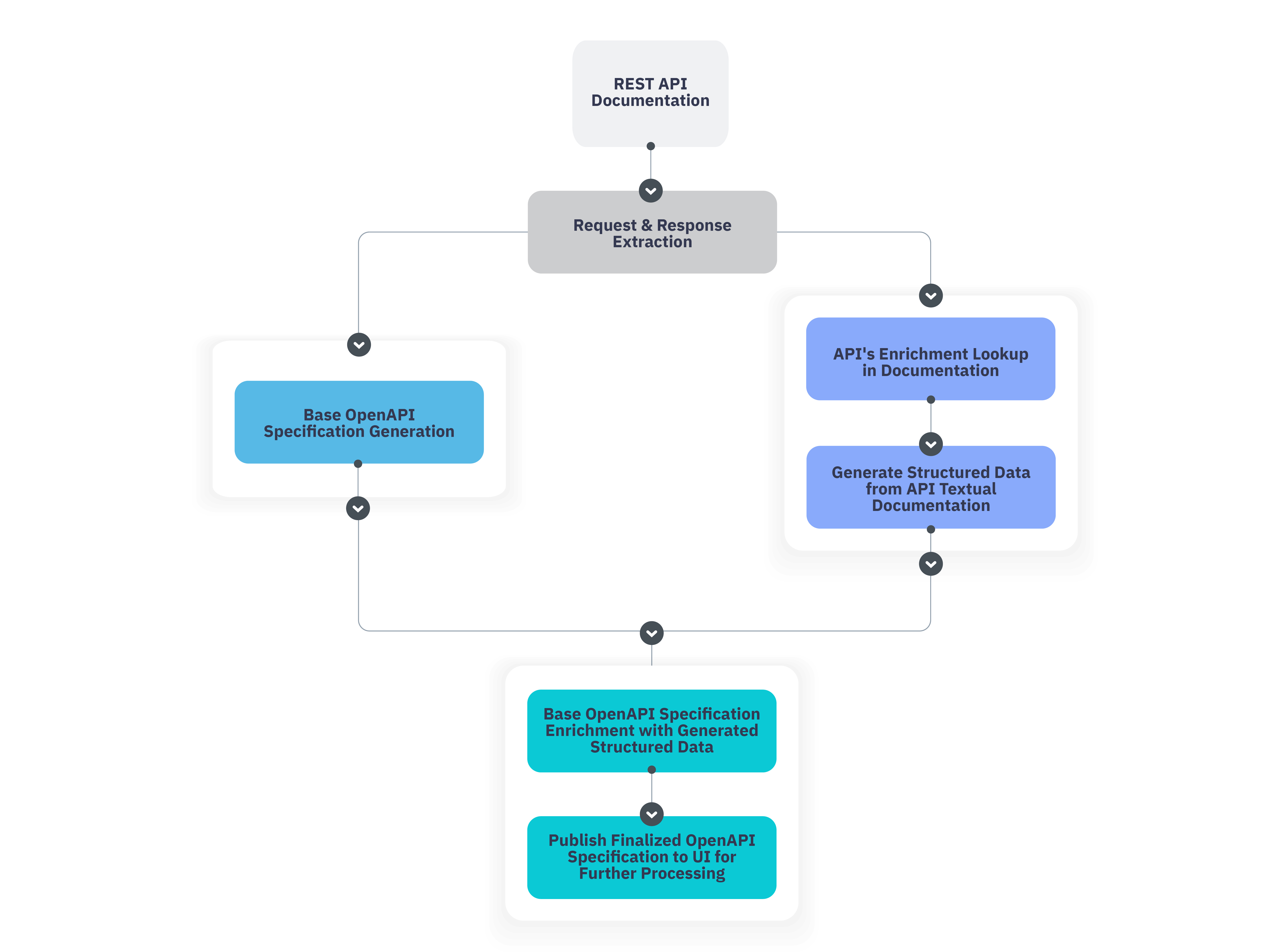}
    \caption{\label{fig:specrawler-system} SpeCrawler Architecture: This diagram demonstrates the carefully designed steps involved in transforming REST API documentation into an accurate OpenAPI specification (OAS). The process begins by extracting pairs of request and response elements from the API documentation webpage, which serve as the foundation for creating a skeletal OAS, as described in details in~\ref{subsec:openapi-spec-generation}. Then, the descriptive section of the documentation is used to gather comprehensive details about the API, its request and response elements, and their parameters, as described in~\ref{subsec:enrichment-specrawler}. Subsequently, the outputs from both processes are integrated to form a comprehensive OAS. Both procedures rely on large language models for generation.
    }
\end{figure*}

Throughout this paper, we explore the multifaceted nature of our staged system methodology, clarifying how each component contributes to a nuanced comprehension of API documentation. Furthermore, we present empirical evidence and case studies to highlight the ability of LLMs to overpower diverse and heterogeneous API documentation structures, producing accurate and comprehensive OASs.
This contribution represents a step forward in improving RESTful API documentation, offering practical benefits for developers and organizations seeking to streamline their integration processes across API-driven systems.

Moreover, due to the vital need to ground LLMs with accurate data, there has been lately a significant surge in the adoption of various tools, i.e. APIs, in the context of LLMs. These tools encompass a spectrum of resources, spanning from custom functions and external APIs such as Wikipedia and the Google Knowledge Graph to knowledge aggregation mechanisms like databases.
Utilizing such tools necessitates the organization of their metadata, including but not limited to their name, definition, inputs, and outputs. 
% However, just like crafting OASs, this manual effort proves to be demanding and labor-intensive. 
Automatically generating OAS for these tools could assist us in developing a wide array of readily available tools for LLMs.

%To summarize, our work makes significant contributions in the following ways:
%\begin{itemize}
%  \item \textbf{Automated OAS Generation:} SpeCrawler generates OAS from diverse API documentation webpages, leveraging LLMs. This significantly reduces manual effort and streamlines the process of creating standardized API documentation from scratch.
  %\item \textbf{Adaptability to Dynamic Landscapes:} SpeCrawler strategically addresses challenges posed by complex API documentation through a pipeline of carefully designed and implemented steps. It effectively manages issues like various elements, incomplete parts, intricate API details, and constantly changing documentation formats, making the documentation process more efficient and accurate.
  %\item \textbf{Innovative Methodology:} SpeCrawler pioneers automated OpenAPI generation by leveraging the power of LLMs and adapting to the dynamic landscape of RESTful API documentation. This represents a significant advancement in the field of API documentation, offering a more adaptable and scalable solution compared to traditional rule-based methods.
  %\item \textbf{Efficiency and Accuracy:} SpeCrawler offers an efficient and accurate approach to API documentation, aiding developers and organizations in refining integration processes within API orchestrating systems. By automating the generation of OAS, it enhances productivity and ensures consistency in API documentation practices.
%\end{itemize}

\section{OpenAPIs and Documentation Websites}

The OpenAPI Specification (OAS) is a widely adopted standard in the realm of API development, offering a formalized structure for describing RESTful APIs. It not only delineates the architecture of APIs but also establishes a comprehensive set of rules governing the definition of API endpoints, request and response parameters, and overall functionality. Furthermore, the use of OAS streamlines the development process by facilitating the automatic generation of client libraries and server-side code, fostering consistency and interoperability across diverse programming environments. Figure~\ref{fig:oas-schema} illustrates an OAS hierarchy.

%On the other hand, a documentation site is an indispensable web-based resource that serves as a central hub for disseminating information about an API. It often goes beyond the mere technical details, encompassing comprehensive documentation on API endpoints, request payloads, response formats, and authentication mechanisms. The inclusion of practical examples and usage scenarios makes it an invaluable tool for developers seeking to understand and implement the API efficiently. A well-designed documentation site is instrumental in reducing the learning curve associated with API adoption, enabling developers to swiftly integrate the API into their projects. As the number of APIs increases, the manual integration of APIs by developers becomes increasingly time-consuming, necessitating the use of automated tools for integration. However, these tools require standardized formats, prompting the need to convert documentation into an API Specification. Moreover, despite the fact that OpenAPI Specifications (OAS) are often utilized to automate the generation of comprehensive documentation, they are not publicly exposed.

Manually creating an OAS requires a lot of human expert effort and meticulous attention to detail, particularly when dealing with a wide range of API documentation formats and sources. Moreover, developers frequently encounter challenges in the creation and upkeep of OAS, leading to potential discrepancies between the APIs deployed in production and their intended specifications \citep{martin2021black, martin2022online, kim2022}. Therefore, automating this process is essential. Yet, it poses various significant challenges. The majority of web API documentations do not follow a machine-readable convention \citep{danielsen2013}. Furthermore, the structural diversity, varied formats, and scattered components across documentation hinder the development of a comprehensive and automated OAS generation solution~\citep{Yang-2018}. Key API components might be scattered throughout the documentation, requiring intelligent extraction. Additionally, critical information may be absent or fragmented, hindering a comprehensive and accurate transformation into OAS. The linguistic variation in the documentation adds complexity to the interpretation and standardization process, while diverse organizational structures within API documentation demand adaptability in the transformation process.

\section{SpeCrawler}
\begin{figure}
    \centering
     \includegraphics[width=.9\linewidth]{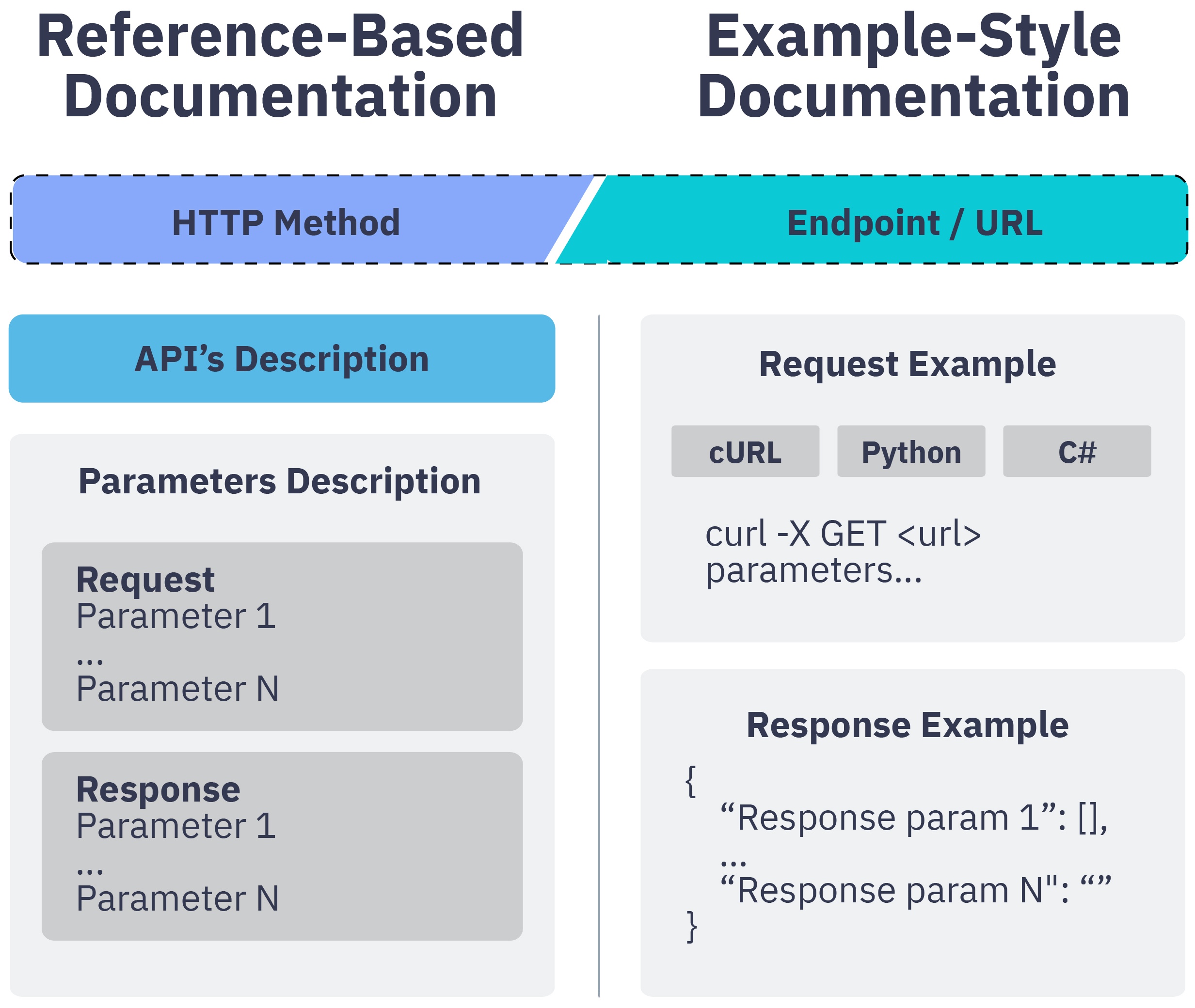}
    \caption{\label{fig:webpage-example} API documentation webpage - This diagram illustrates a typical structure found in online API documentation webpages, including its key components. On the left side, you'll find the reference-based documentation, which primarily comprises descriptive text explaining the API, its request and response elements, parameters, and additional metadata. On the right side, the example-style documentation section provides practical demonstrations of API interaction, including common request examples, and sample responses users can anticipate. The API's HTTP method and URL is commonly featured on either side.
    % A dashed cyan line separates the reference-based (left) and the example-style (right) documentation. A green rectangle marks elements from which we derive the input to the LLM. Indications for matching between the generated OpenAPI specification and the current HTML scope are marked by a red rectangle.   
    }
\end{figure}

% In this section, we introduce SpeCrawler, an innovative multi-staged approach designed to autonomously produce OpenAPI Specifications from API documentation websites. Firstly, detailed in Section~\ref{subsec:scraping}, we outline our initial scraping stage where we gather API request and response pairs. This stage serves the dual purpose of providing input for subsequent stages and facilitating the parallelization of generation tasks. Secondly, expounded upon in Section~\ref{subsec:openapi-spec-generation}, we delve into the process of generating an initial OpenAPI Specification (OAS) using a Large Language Model (LLM) based on the collected request and response pairs. Lastly, in Section~\ref{subsec:enrichment-specrawler}, we elucidate the generation of structured data concerning API parameters by an LLM, utilizing the request and response pairs, and describe the integration of this data into the initially generated OAS.
% The illustration provided in Figure \ref{fig:specrawler-system} delineates the various stages involved in crafting the SpeCrawler system.

 In this section, we present SpeCrawler, a new system designed to autonomously create OpenAPI Specifications from API documentation websites. First, we describe the initial scraping stage in Section~\ref{subsec:scraping}, where we identify all API request and response example pairs from a documentation page. This stage provides the necessary components for the subsequent two stages. Second, in Section~\ref{subsec:openapi-spec-generation}, we discuss the process of employing a generative LLM to create a base OAS given the request and response example pairs. Lastly, in Section~\ref{subsec:enrichment-specrawler}, we explain how we extract relevant descriptions about the API parameters and generate structured data for enriching the base OAS. Figure \ref{fig:specrawler-system} illustrates the different stages of developing the SpeCrawler system.

\subsection{Scraping}
\label{subsec:scraping}
The first step in scraping the API documentation page involves retrieving example-style documentations, which consist of usage examples demonstrating the requests and responses of the APIs documented. An illustration of an example-style documentation is depicted on the right side of Figure~\ref{fig:webpage-example}. To identify and extract the specified elements, we developed a scraper that searches for HTML elements within the webpage containing example-style components like cURL commands and JSON responses, following predefined rules and patterns. Since some HTML elements are loaded dynamically, the scraper also navigates through the documentation page, activating various buttons using heuristic techniques to fetch such relevant elements during runtime. 
% Additionally, it scans the page to identify API requests and response examples through rule-based heuristics.
% If cURL commands are not found, several alternatives such as These elements can be identified and extracted with high accuracy with heuristics based on pattern-matching techniques. 
% The scraper also searches for tables of parameters, which are often included in API documentation. These tables contain valuable information about the input and output formats of the API endpoints.

% Once the relevant elements have been extracted, the next step is to align the requests and the responses. To that end, we leverage the HTML structure of the webpage to implement a heuristic that traverses the HTML DOM Tree to find the closest response available given a request example. 

After identifying and extracting the relevant components, we aim to align between the different request and response elements, as these pages may contain multiple APIs. For this purpose, we utilize the HTML structure of the webpage to apply a heuristic that navigates the HTML DOM Tree to locate the nearest available response corresponding to a given request example.
Upon completing this process, we possess a comprehensive list of all APIs represented as pairs of request and response examples. 
% This enables us to proceed to the step of generating the base OAS.

% This can be achieved using a positional heuristic, which takes into account the position of the elements in the HTML code. By analyzing the position of the CURL commands and the responses, we can determine which response corresponds to which command. 
% This allows us to create a mapping between the CURL commands and the endpoint responses, which is essential for generating accurate open API specs.

\begin{figure*}
    \centering
     \includegraphics[width=.9\linewidth]{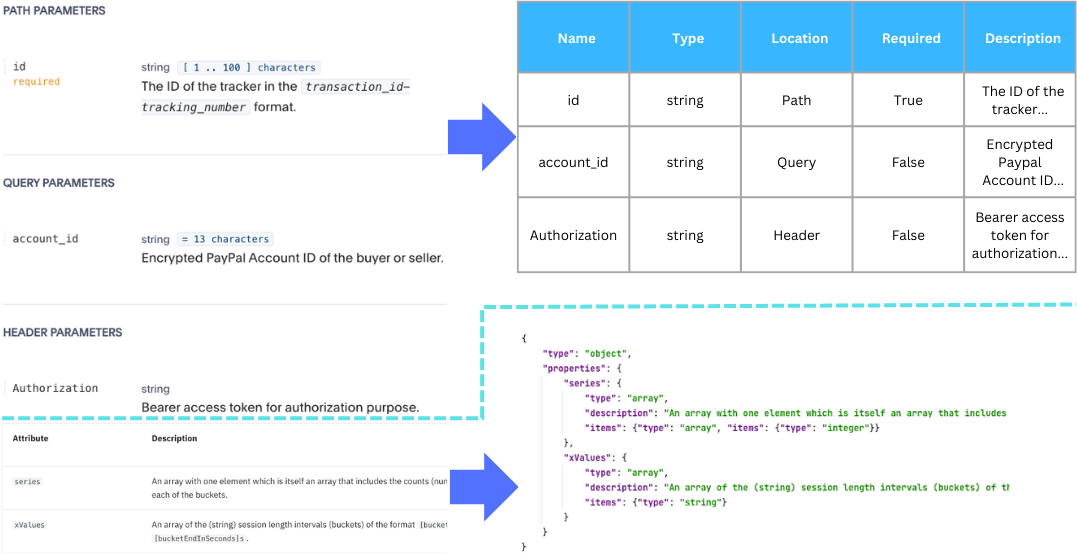}
    \caption{\label{fig:request-enrichment-example} Labeled data - This figure provides a visual representation of input-output pairs from the enrichment generation stage, sourced from PayPal Developer and Amplitude APIs. The top section displays examples of a request element, while the bottom section displays a response element example. On the left side of the figure, you'll find input sources, which consist of scraped raw HTML scopes from API documentation websites. On the right side, the results of the enrichment generation process are presented. For request elements, the results are formatted as a TSV table, while for response elements, they are showcased as a response OpenAPI schema nested object.
    }

    %This figure displays an example of input-output pairs from the request (top) and response (bottom) enrichment generation stage, sourced from PayPal Developer and Amplitude APIs. The left side displays input sources, comprising scraped raw HTML scopes from API documentation websites. On the right side, the results of the enrichment generation process are presented as a TSV table for the request and a response OpenAPI schema object for the response.

    % \footnote{\url{https://developer.paypal.com/docs/api/tracking/v1/#trackers_get}}
    % A dashed cyan line separates the reference-based (left) and the example-style (right) documentation. A green rectangle marks elements from which we derive the input to the LLM. Indications for matching between the generated OpenAPI specification and the current HTML scope are marked by a red rectangle.   
\end{figure*}

\subsection{Base OAS Generation}
\label{subsec:openapi-spec-generation}

% Developing an OAS involves navigating the documentation webpage's various intricate elements and factors. The diverse and intricate nature of API documentation pages poses challenges for employing a rule-based method effectively to address all possible webpages. As a result, we opted to leverage LLMs as a solution. 
Having obtained pairs of request and response examples in the preceding section, we would like to generate a base OAS. This can be accomplished through a complex rule-based algorithm which automatically parse each pair and construct the OAS accordingly. However, this method is prone to vulnerabilities arising from potential human errors during the composition of these examples, as well as noise introduced by the automated scraping process. 

Another approach is utilizing LLMs that demonstrate proficiency in extracting and organizing information into a structured format even when faced with various source formats, especially when they have been trained with similar structures beforehand.

Yet, in a preliminary experiment, we observed that expecting the LLM to generate the entire OAS in a single prediction might pose significant challenges for the model, even with the most sophisticated hand-crafted prompt. Therefore, we opted to break down the task into more manageable subtasks.
The first subtask involves generating a skeleton structure of the OAS based on the request examples found in the scraping. This skeleton serves as a foundational framework for the next generation phase. Crucial metadata about the OAS is embedded within the generated skeleton. This includes information such as the servers, outlining the locations where the API is hosted, the specified method for communication, security protocols employed, and details regarding header and path parameters. 
In the next subtask, we generate JSON schemas from the request and response examples obtained. This involves extrapolating structural and semantic information from the examples to construct a comprehensive and standardized representation as a JSON schema. Understanding the potential intricacies of request and response objects, especially when they are excessively long or nested, led us to adopt a methodology that involves breaking down these objects into multiple segments based on a specified line  threshold. Fragmenting the request or response examples into more manageable components ensures the model can efficiently analyze and remember all parameters. After the model processes these small segments, we aggregate them and embed them in the skeleton OAS.

\subsection{Enrichment}

\label{subsec:enrichment-specrawler}

API documentation webpages typically feature example-style documentation, such as cURL commands and their corresponding responses, as well as reference-based API documentation, clarifying the API's objectives and parameters. An illustration of such a webpage is depicted in Figure ~\ref{fig:webpage-example}. 

Enriching the base OAS outlined in the preceding section with reference-base documentation is imperative because the example-style documentation lacks much of the information found in reference-based documentation.

% While the example-style documentation provides indispensable information for generating an OAS,  explained in the previous section, the utilization of an LLM is feasible for generating an OpenAPI Specification from an example-style documentation. Unfortunately, such generated specification does not encompass the essential information contained in the reference-based documentation. Therefore, enriching the OpenAPI is imperative. 
Employing deterministic rule-based algorithms to accurately parse of these documentations is impractical due to the substantial differences in HTML structures across various websites \citep{Yang-2018}. For example, in a preliminary experiment, we tried to parse HTML parameter tables with pandas \textit{read\_html} function\footnote{\url{https://pandas.pydata.org/docs/reference/api/pandas.read_html.html}}, yet it repeatedly failed to correctly identify the columns and rows of the table. Therefore, turning to LLMs emerges as a viable solution, given their ability to generalize over such structural differences.
Nevertheless, it is important to recognize that LLMs possess constraints in terms of their context length. Consequently, filtering the documentation becomes imperative prior to the utilization of the LLM.

Therefore, we developed an algorithm to automatically find the essential HTML elements from the API documentation webpage that encapsulate the crucial information needed for enhancing the base OAS. Since the HTML structure of API documentation websites may vary significantly, the algorithm mainly relies on semantic signals rather than specific HTML syntax or structure. After finding the minimal scope, we further preprocess and filter it to optimize the signal-to-noise ratio. We explain the algorithm thoroughly in Appendix~\ref{subsec:appendix-find-minimal-ancestor}. 
% Then, given the preprocessed HTML, the LLM generates a structured data representation by in-context learning, facilitating the automatic merge of the generated base OAS from the preceding section.

Having obtained the processed minimal ancestor, we employ an LLM to generate structured data for enhancing the base OAS. The LLM is directed to produce a TSV table for the request and an OpenAPI response schema for the response. We chose the TSV format for the request, as documentation for request parameters typically lacks a hierarchical structure compared to response documentation. Examples of the inputs and outputs of both enrichments are provided in Figure~\ref{fig:request-enrichment-example}.
For each parameter, the model is instructed to generate the name, type (e.g., string), whether the parameter is required, and a description if present. For request parameters specifically, we additionally prompt the model to generate the location (e.g., query, body). 
Opting for in-context learning (ICL) during LLM generation was our preference, as it eliminates the need for extensive labeling, such as in supervised fine-tuning. To that end, we manually labeled 15 examples from different websites. From these, we selected 3 in-context examples for each input by computing the cosine similarity of the HTML tags' frequency distributions between each example and the given input for prediction \footnote{we also experimented with KL divergence as a similarity metric.}. The selection of HTML tag frequency distributions is motivated by their ability to approximate the similarity in HTML structures, thereby offering the LLM parsing examples that closely resemble those it is tasked with performing. A representative prompt example is elucidated in Appendix~\ref{subsec:appendix-prompt-generation-examples}.
Following the data generation, we validated the structure of the output data (TSV/JSON) and removed any hallucinations by verifying if the parameter names explicitly appeared in the input.

Finally, we incorporated the generated structure into the base OAS in a rule-based fashion. This integration prioritizes the description and required parameter fields from the generated enrichment, while giving precedence to the type and location fields from the base OAS. The basis for this prioritization lies in the types of documentation used as input for generating the two components.

\section{Experiments}
In this section we present the results of a series of experiments conducted to evaluate the performance of SpeCrawler. Specifically, we compare the performance of various LLMs on the task of a base OAS generation, followed by an examination of their performance on the task of OAS enrichment given request and response examples. Finally, we conduct an end-to-end evaluation of SpeCrawler and compare its performance to that of other approaches and models.

\subsection{Base OAS Generation}
\label{subsec:results-base-oas-generation}
In this section we analyzed the syntactic properties of automatically generating base OASs using LLMs given a corpus of $49$ documentation pages, collectively encompassing approximately 189 endpoints. We chose several renowned LLMs for our experiment, which exhibited in-context prompting capabilities: llama2-70b-chat~\citep{touvron2023llama}, codellama-34b-instruct~\citep{rozière2024code}, mistral-7b-instruct~\citep{jiang2023mistral}, mixtral-8x7b-instruct~\citep{jiang2024mixtral} and IBM's Granite model\footnote{\url{https://www.ibm.com/downloads/cas/X9W4O6BM}}. 
% The granite-20b-code-instruct-v2 model represents an instruction-tuned iteration, initiated from the granite-20b-code-v2 base model, which was pre-trained on diverse code corpora. This model has undergone supervised fine-tuning, guided by the work of \citet{wei2021finetuned}, aimed at enhancing its ability to follow instructions effectively. This refinement facilitates the model's utility in completing enterprise tasks through prompt engineering. 
Granite is part of IBM Generative AI Large Language Foundation Models, which are Enterprise-level English-language models trained with large a volume of data that has been subjected to intensive pre-processing and careful analysis.
We utilized the Granite model, along with the other LLMs, also in Sections~\ref{subsec:enrichment-results},~\ref{subsec:end-to-end-results}.

The syntactic evaluation was centered on three key measures: (1) whether the generated output conforms to the JSON syntax standards; (2) whether the generated output conforms to the OAS format; (3) average number of warnings in the generated valid JSONs. Warnings appear in scenarios where the generated specification is a valid JSON but fails to meet the criteria of a valid OAS. These warnings signify potential inconsistencies or deviations from the OAS standard. The average number of warnings per case was calculated to provide insights into the degree of syntactic divergence from the standard. We evaluated the last two measures using the jsochema library, a widely-used tool for validating OpenAPI and JSON specifications~\footnote{\url{https://github.com/python-jsonschema/jsonschema}}.

The findings derived from the syntactic analysis are outlined in Table~\ref{tab:Generation_Results}. Notably, the Granite and Code Llama models~\citep{rozière2024code} emerge as the top-performing models. It is noteworthy that while Code Llama attained the highest proportion of valid OpenAPI Specifications (OAS), its warnings ratio ranked third when compared to the other models. Conversely, the Granite model secured the second position in the ratio of valid OAS but demonstrated the lowest incidence of warnings in the valid JSON outputs. Apart from these two models, the remaining models generally succeeded in producing valid JSONs in the majority of cases. However, their rates of generating valid OAS were notably low and exhibited significant variability. Hence, it can be inferred that even when decomposing the OAS generation into subtasks, it remains a nontrivial challenge for LLMs.

\begin{table}[t]
\caption{Syntactic evaluation results for base OAS generation applying different LLMs.}
\label{tab:Generation_Results}
\begin{center}
\begin{small}
    
\begin{sc}
\begin{tabular}{@{}llll@{}}
\toprule
                     & Valid Json          & Valid OAS       & Warnings   \\ \midrule
granite      &  \boldsymbol{$1$} &  $.73$     &  \boldsymbol{$.48$}  \\
llama-70b    &  \boldsymbol{$1$}  &  $.29$    &  $.78$  \\
codellama    &  $.99$     &  \boldsymbol{$.89$}  &  $.59$ \\
mistral      &  \boldsymbol{$1$}     &  $.4$  &  $.54$ \\
mixtral      &  $.92$   &  $.66$ &  $.64$ \\ \bottomrule
\end{tabular}
\end{sc}
\end{small}
\end{center}

\end{table}

\subsection{Enrichment Generation}
\label{subsec:enrichment-results}
% \begin{table}
% \centering
% \begin{tabular}{@{}lcccc@{}}
% \toprule
% & F1  & Req. & Type & Loc. \\ \midrule
% granite-20B & .94 & .91  & .75 & .9 \\
% codellama-34b & .94 & .91 & .72 & .8 \\
% GPT-4 & 1 & .98 & .64 & .9 \\ \bottomrule
% \end{tabular}
% \caption{Performance Comparison of Generative LLMs on Enrichment Generation for API Parameters.}
% \label{tab:enrichment}
% \end{table}

% \usepackage{booktabs}
\begin{table*}[t]
\caption{Assessment of the Enrichment Generation Stage. F1 score is calculated based on the parameter names correctly identified in the input. Precision scores for the required (Req.), type (Type), and location (Loc.) parameter fields, as well as the description (Desc.) similarity are computed over accurately generated parameters.}
\label{tab:enrichment}
\begin{sc}
\begin{center}
\begin{small}
\begin{tabular}{@{}l|lllll|llll@{}}
\toprule
\multicolumn{1}{c|}{}     & \multicolumn{5}{c|}{Request Enrichment}                                 & \multicolumn{4}{c}{Response Enrichment}                  \\ \midrule
                          & F1           & Req.         & Type         & Loc.        & Desc.        & F1           & Req.         & Type         & Desc.       \\ \midrule
Granite & $.94$          & $.91$          & \boldsymbol{$.75$} & \boldsymbol{$.9$} & $.87$          & $.88$          & $.67$          & $.91$          & $.83$         \\
Llama2           & \boldsymbol{$.99$} & \boldsymbol{$.98$} & $.72$          & $.79$         & $.84$          & $.85$          & $.5$           & $.91$          & \boldsymbol{$.9$} \\
Codellama    & $.94$          & $.91$          & $.73$          & $.8$          & $.87$          & \boldsymbol{$.97$} & $.66$          & $.92$ & $.89$         \\
Mistral       & $.96$          & $.88$          & $.65$          & $.82$         & $.84$          & $.96$          & \boldsymbol{$.77$} & $.89$          & $.66$         \\
Mixtral     & \boldsymbol{$.99$} & $.82$          & $.72$          & $.84$         & \boldsymbol{$.91$} & $.76$          & $.59$          & \boldsymbol{$.94$}          & $.88$         \\ \bottomrule
\end{tabular}
\end{small}
\end{center}
\end{sc}
\end{table*}
In order to assess the efficacy of the enrichment generation in isolation, we carried out a constrained leave-one-out cross-validation trial. Here, $n-1$ manually labeled examples served as potential in-context instances from which we choose $3$ in-context examples, while one example was set aside for testing purposes. A short prompt example is provided in Appendix~\ref{subsec:appendix-prompt-generation-examples}. We tagged a total of $n=15$ challenging examples from different API documentation websites. We devised and experimented with different generation prompts, and the formulation that underwent testing and usage was optimized through the falcon-40b model~\citep{falcon40b}. First, we assessed the model's capability to accurately extract the parameter names of the API using the F1 score, which accounts for both hallucinations and deficiencies in recall abilities. Next, we assessed the precision scores associated with the parameter fields: "required," "type," and "location," ensuring that the model extracted and generated them accurately, as occasionally the model may omit or hallucinate different values in these fields. For evaluating the extraction and generation of the "description" field, we utilized cosine similarity between the ground-truth description and the generated one. This approach is preferred over an exact-match style score, as it might be overly stringent. For inputs for the LLM, we took manually-crafted HTML scopes.

Table~\ref{tab:enrichment} compares several LLMs on the aforementioned metrics in both request and response enrichment separately. We note several interesting observations based on these results.
% According to the findings presented in  Table~\ref{tab:Generation_Results}, it is evident that 
LLMs successfully retrieved a majority of the parameter names.
The LLMs achieved a high F1 score in both request and response enrichments, thereby showcasing the capability of such models to generalize across diverse HTML structures.

Response enrichment is harder.
The scores of the response enrichment are lower on average, except for the type precision. This can be explained by several factors. First, the response object is often highly-nested which is harder to generate compared to the flat TSV structure of the request enrichment. Second, the information about the ``required'' fields has to be generated as a separate list for each scope for the response. Third, the response often has no enrichment which may lead to hallucinations. This stands in contrast to request parameters, which tend to consistently appear and vary among APIs. The exceptional type precision is higher for the response possibly due to its versatility or absence in reference-based documentation of the request.

Output format matters for the model.
Several models demonstrated significantly enhanced performance in one type of enrichment type compared to the other, suggesting the importance of customizing the output format to suit individual models. For instance, Code Llama and Mistral, which attained the highest F1 scores for response enrichment, did not exhibit exceptional performance in request scenarios. Conversely, Llama2 and Mixtral, which achieved the highest F1 scores for request enrichment, performed poorly in response scenarios, registering the lowest F1 scores.

\subsection{End-to-End Testing}
\label{subsec:end-to-end-results}
\begin{table*}[t]
\caption{End-to-end results between models with and without enrichment. Precision (P), Recall (R), F1-score (F1), and Similarity (Sim) metrics are reported for both request and response phases. With enrichment, models exhibit varied performance improvements across metrics, with the Granite model showing significant enhancement in Precision, F1-score, and Similarity.}
\label{tab:e2e}
\begin{center}
\begin{small}
\begin{sc}
\begin{tabular}{@{}lllllllll|llllllll@{}}
\toprule
\multicolumn{1}{c}{} & \multicolumn{8}{c}{With Enrichment}                                  & \multicolumn{8}{c}{Without Enrichment}   \\ \midrule
\multicolumn{1}{c}{} & \multicolumn{4}{c}{Request}                                  & \multicolumn{4}{c}{Response}  & \multicolumn{4}{c}{Request}        & \multicolumn{4}{c}{Response}                         \\ \midrule
                     & P          & R       & F1 & Sim        & P    & R  & F1     & Sim     & P          & R       & F1    & Sim   & P          & R       & F1    & Sim     \\ \midrule

        granite & \boldsymbol{$.74$} & \boldsymbol{$.46$} & $.51$ & $.38$ & $.96$ & \boldsymbol{$.46$} & $.72$ & \boldsymbol{$.85$} & \boldsymbol{$.72$} & $.42$ & $.47$ & $.35$ & \boldsymbol{$.97$} & $.42$ & $.66$ & \boldsymbol{$.88$} \\ 
        llama-70b & $.37$ & $.32$ & $.61$ & $.28$ & $.95$ & $.32$ & $.49$ & $.72$ & $.37$ & $.32$ & $.61$ & $.28$ & $.95$ & $.32$ & $.49$ & $.72$ \\ 
        codellama & $.62$ & $.44$ & $.58$ & \boldsymbol{$.50$} & $.95$ & $.44$ & $.64$ & $.80$ & $.62$ & $.44$ & $.58$ & \boldsymbol{$.50$} & $.95$ & $.44$ & $.64$ & $.80$ \\ 
        mistral & $.24$ & $.25$ & \boldsymbol{$.67$} & $.22$ & $.95$ & $.25$ & \boldsymbol{$.85$} & $.39$ & $.24$ & $.25$ & \boldsymbol{$.67$} & $.22$ & $.95$ & $.25$ & \boldsymbol{$.85$} & $.39$ \\ 
        mixtral & $.41$ & $.38$ & $.60$ & $.25$ & \boldsymbol{$.97$} & $.38$ & $.55$ & $.57$ & $.53$ & \boldsymbol{$.46$} & $.60$ & $.25$ & $.93$ & \boldsymbol{$.46$} & $.61$ & $.79$ \\ \hline

\end{tabular}
\end{sc}
\end{small}
\end{center}
\end{table*}

In this section, our emphasis lies in assessing the overall capabilities of the SpeCrawler system. We conducted experiments to compare the system's performance while using different generative LLMs. 
To that end, we employed a manually curated dataset comprising 30 API documentation pages, each linked to a corresponding ground-truth OAS. The dataset encompasses approximately one hundred API endpoints. 
In each experiment we used the in-context learning approach with the same prompt and the same three in-context examples.
% In this experiment, we collected 30 documentation pages along with their corresponding OAS. 
Each generated specification was compared against the ground-truth OAS. Therefore, for the system to perfectly match the ground-truth OAS, it needs to retrieve precisely all the parameters in the request, even if they do not appear in the example-style documentation, i.e. in the request or response usage examples.
The outcomes are detailed in Table~\ref{tab:e2e}, which displays metrics such as Precision (P), Recall (R), F1 score (F1), and cosine similarity (Sim) for both request and response aspects, with and without enrichment, as discussed in Section ~\ref{subsec:enrichment-specrawler}.

% The evaluation metrics presented offer valuable insights into the system's efficacy in precisely extracting and generating parameters from both the request and response components. 
% The precision score assesses the accuracy of the generated parameters, while the recall score evaluates the system's capability to identify and extract all relevant parameters. The F1 score combines these metrics to provide a balanced evaluation. Additionally, the non-exact-match cosine similarity metric approximates the degree to which the extracted descriptions align semantically with the ground-truth ones.

The results presented in Table~\ref{tab:e2e} illustrate the system's performance under various conditions, taking into account the inclusion or exclusion of the enrichment component. 
Firstly, activating the enrichment component results in a notable improvement in precision, recall, and F1 scores for both request and response components. Nevertheless, the improvement in the similarity metric fluctuates depending on the choice of the base model; for example, employing IBM's Granite model results in a 4-point increase. This observation can be attributed to the propensity of LLMs to generate descriptions for parameters even in the absence of such descriptions in the input during the enrichment generation stage.
Secondly, when compared to the results presented in Sections~\ref{subsec:results-base-oas-generation} and \ref{subsec:enrichment-results}, which evaluated the performance of particular generation components, the models exhibit significantly lower performance in this context. This emphasizes the intricacy of the task and underscores the continued necessity for improvements in rule-based components such as scraping and determining the minimal HTML scope for enrichment. 
Thirdly, upon scrutinizing the dataset, we identified various instances of noise in the annotation, suggesting that some of the performance degradation can be attributed to this factor as well. 
Finally, it is noteworthy that IBM's Granite demonstrated commendable outcomes across various metrics, achieving the best performance overall, particularly excelling in precision and recall of parameter names. This observation is somewhat unexpected, given the prominence of its competitors, and that neither the prompts nor the hyperparameters were customized for this model.

\subsection{Comparing Against LLM-Based Solutions}

\begin{table*}[t]
\caption{End-to-end comparison against other LLM-based solutions on syntactic measures of OAS (see Section~\ref{subsec:results-base-oas-generation}) and semantic measures (precision and recall of parameter names).}
\label{tab:openai}
\begin{center}
\begin{small}
\begin{sc}
\begin{tabular}{@{}lcccccr@{}}
\toprule
Measure & GPT4-Turbo & ActionGPT & SpeCrawler \\
\toprule

Valid Json & $.4$ & \boldsymbol{$1$} & \boldsymbol{$1$}  \\
Valid OAS & $.4$ & \boldsymbol{$1$} & \boldsymbol{$1$}  \\
Warnings & \boldsymbol{$0$} & \boldsymbol{$0$} & \boldsymbol{$0$}  \\
Precision & $.18$ & $.53$ & \boldsymbol{$.54$}  \\
Recall & $.05$ & $.31$ & \boldsymbol{$.47$}    \\
%Nesting accuracy & $.45$ & $.51$ & \boldsymbol{$.66$} \\
\end{tabular}
\end{sc}
\end{small}
\end{center}
\end{table*}

In this series of experiments, we conducted a comparative analysis among three potential solutions for OAS generation from online documentation pages: GPT4-Turbo, OpenAI's latest-generation model incorporating a 128k context window, ActionGPT, a solution proposed by OpenAI for generating documentation from URLs, and our SpeCrawler system with IBM's Granite generative LLM, which was introduced in Section~\ref{subsec:results-base-oas-generation}. We did not compare against rule-based approaches as  we were unable to locate existing comparable results for such methods.
 
The evaluation involved generating an OAS from $10$ manually picked challenging documentation URLs. While some pages contained multiple endpoints, the assessment focused on a single specific selected endpoint. Precision and recall scores were measured in terms of the OAS parameter names generated by each approach, along with three metrics introduced in Section~\ref{subsec:results-base-oas-generation} that measure the structural validity of the generated outputs.

% The models were evaluated based on their performance on these metrics: precision, recall, valid JSON, valid OAS, warnings.

The evaluation results are presented in Table~\ref{tab:openai}. Firstly, SpeCrawler demonstrated the best performance across all metrics compared to the other solutions. Since the Granite model was not fine-tuned or aligned to this specific task, we attribute SpeCrawler's main advantage to its carefully designed pipeline, which effectively decomposes this challenging task into more manageable subtasks.
Secondly, GPT4-Turbo exhibited lower performance across all metrics, with a precision score of $0.18$, recall score of $0.05$, and valid OAS ratio of $0.4$. These results can be ascribed to the HTML content of six URLs being excessively lengthy, exceeding the substantial context window capacity of GPT4-Turbo. This underscores the limitation of a simplistic approach in addressing the task across diverse websites, emphasizing the necessity of a more sophisticated strategy, such as our scraper and enrichment algorithms, to focus the model on relevant contexts, in small bites.
Finally, with regard to warnings, all three models did not exhibit any warnings in their generated valid JSONs, indicating a proficient understanding of the desired schema structure.

The findings of this investigation imply that SpeCrawler represents an advantageous methodology for generating OpenAPI Specifications from complex API documentation websites. Its elevated precision, recall, and valid OAS performance suggest a substantial potential to significantly reduce the time needed for the manual creation of OAS.

%\begin{table}[t]
%\label{tab:e2e}
%\begin{center}
%\begin{small}
%\begin{sc}
%\begin{tabular}{@{}lllll@{}}
%\toprule

%                     & P          & R       & F1 & sim  \\ \midrule
%granite      &  $.86$ &  $.54$     &  $.73$  & $.36$ \\
%llama-70b    &  $.79$  &  $.39$    &  $.57$ &  $.21$ \\
%codellama    &  $.90$     &  $.51$  &  $.71$ & $.11$\\
%mistral      &  $.87$     &  $.54$  &  $.79$ & $.2$ \\
%mixtral      &  $.87$   &  $.31$ &  $.48$ &   $.18$\\ \bottomrule
%\end{tabular}
%\end{sc}
%\end{small}
%\end{center}
%\end{table}

\section{Related Work}
Historically, varied methods have been adopted to generate OASs from API documentation. AutoREST \cite{cao2017automated} generates an OAS by finding all relevant linked HTML documentation pages within the same domain as the root page of the REST API documentation. The OAS is then generated based on information extracted through a set of fixed rules. 
% based on patterns found in popular web services and necessary components such as Base URL, Path Templates, Verbs, and Parameters are extracted using distinct strategies.
D2Spec \cite{yang2018automatically} aims to extract base URLs, path templates, and HTTP method types, using rule-based web crawling techniques and classic machine learning to identify potential API call patterns in URLs. It also analyzes URL paths hierarchically to identify and group path parameters.
Respector~\cite{huang2024generating} employs static and symbolic program analysis to automatically generate OAS for REST APIs from their implementations. 
SpyREST \citep{sohan2015spyrest} employs an HTTP proxy server to intercept HTTP traffic to generate API documentation.
% In recent work, Respector \cite{huang2024generating} focuses primarily on annotation-based REST API frameworks and generates a database of patterns from these framework annotations and library methods. Planting its feet firmly in the source code of the API, it identifies various elements and generates endpoint method specifications using specific algorithms. Respector also simplifies path constraints and detects parameters within request bodies, before constructing an OpenAPI 3.0 compliant API specification.
% In a study by \cite{bahrami2020automated}, a parallel web crawler is utilized to gather diverse API documentations. These are then sorted into REST or non-REST using the REST API Filter, a logistic regression model. To generate the API specification, the method applies regular expression patterns for tasks like API endpoint extraction and utilizes an API Language Model to enhance pattern recognition across various API documentations.
\citet{bahrami2020automated, bahrami2020deep} combines rule-based and machine-learning algorithms to generate OAS from API documentation. They also develop a deep model to pinpoint fine-grained mapping of extracted API attributes to OAS objects. WATAPI \citep{bahrami2019watapi} takes a different approach by adding a user as a human-in-the-loop to interact transparently with complex machine-learning components to compose an OAS.
SpeCrawler system presents a novel and unified approach by leveraging LLMs to automate the generation process directly from raw HTML content, enabling a more robust and accurate solution across diverse API documentation structures without being bound to specific patterns and custom mechanisms.
Most similar to our work, \citet{androvcec2023using} used GPT-3 to automatically generate OAS from a preprocessed HTML file describing an API documentation. SpeCrawler distinguishes itself from their methodology through two key strategies: (1) dividing the generation task into multiple parts, and (2) extracting relevant information from webpages, thus accommodating webpages that exceeds the context size constraint of GPT-3 and those featuring multiple operations. 

% Several tools for generating OAS were published in recent years, however, they all require the user to perform additional steps in order to generate an OAS \citep{}
% Lastly, SpyREST  \cite{sohan2015spyrest} automates most of the conventional manual process of documenting RESTful APIs, becoming a distinct tool for the task. As opposed to other methodologies, it employs an HTTP proxy server to intercept API calls and captures the resulting HTTP traffic. The data is then structured into a hierarchical model that sequentially classifies API hosts, versions, resources, and actions and generates customizable documentation.

\section{Conclusions}
This paper introduces SpeCrawler, an innovative multi-stage methodology designed to automatically generate comprehensive OAS from online API documentation. Combining rule-based algorithms and generative LLMs, SpeCrawler addresses existing limitations associated with LLMs while showcasing robustness and generalization capabilities across varied HTML structures in API documentation.
The effectiveness of SpeCrawler is demonstrated through satisfactory results across a diverse list of API documentation websites. Therefore, it significantly reduces the manual efforts required by technical experts in crafting OAS manually.

Potential routes for future research include enhancing the extraction of elements utilized for enriching both the request and response elements by enabling multiple HTML scopes as input to the generative model and aggregating their product. Additionally, there is room for improvement in the scraping process by gathering more relevant information from associated links within the documentation. Further exploration could involve investigating various strategies for splitting the generation process, such as consolidating generations for the request and the response elements separately. Another possible direction is enlarging the existing resources to a large-scale dataset to constitute a benchmark for this task.

% \section{Appendices}

% If your work needs an appendix, add it before the
% ``\verb|\end{document}|'' command at the conclusion of your source
% document.

% Start the appendix with the ``\verb|appendix|'' command:
% \begin{verbatim}
%   \appendix
% \end{verbatim}
% and note that in the appendix, sections are lettered, not
% numbered. This document has two appendices, demonstrating the section
% and subsection identification method.



\appendix
\section{Appendix}
\subsection{Find Minimal Ancestor Algorithm}
\label{subsec:appendix-find-minimal-ancestor}
To determine the appropriate HTML scope for enrichment given a request example, we employed two distinct approaches. In scenarios where a webpage incorporates multiple API calls, we defined the scope as the highest ancestor of the request example HTML element that does not encompass other requests~\footnote{If the minimal ancestor of the subsequent request is not consecutive, it is defined as a sequence of elements ending in the ancestor of the next request}. In Figure~\ref{fig:webpage-example}, this scope should encompass both reference-based and example-style sections. Conversely, when dealing with a webpage containing a single API call, we conducted a search for \textit{leaf elements.}\footnote{HTML elements lacking children} likely associated with parameters in the reference-based documentation based on their text, such as parameters from the request or response, and parameter header templates. These elements could be situated, for instance, in the ``Parameters Description'' section as illustrated in Figure~\ref{fig:webpage-example}. Subsequently, we iterated through the ancestors of each identified element, starting from the immediate parent and moving upwards, in search of the first ancestor containing a matching URL endpoint corresponding to the provided API URL. Since this is often found preceding the HTTP method (e.g. ``GET /info/{id}''), we denote it as ``HTTP Method'' and ``Endpoint/URL'' in Figure~\ref{fig:webpage-example}.

% We mark indications for a matching in a red rectangle in Figure~\ref{fig:webpage-example}. 
After retrieving these minimal ancestors, we rank them according to two criteria: 1) the number of parameters from the request or response found as leaf elements in the ancestor, and 2)
% (2) whether an endpoint matching the API URL was found; 
whether the HTTP method type of the URL was found as a leaf element. 
% In Figure~\ref{fig:webpage-example} the HTML scope which encompasses both sides have the parameter `accept` from the request (marked by a green rectangle), an endpoint matching the API URL (marked by a red rectangle), and a leaf element of the `GET` method (marked by a blue rectangle). 
Following this ranking, we filter out HTML elements that are ancestors of other candidates. 
Lastly, if we still have multiple candidates sharing the same rank, we randomly sample one of them, although we did not encounter such cases in our experiments.

The minimal ancestor is then preprocessed to remove noise and tailor it to the constrained context size of the LLM. This involves filtering out its children that are less likely to contain relevant information for augmenting the base OAS. Specifically, we search for parameter names extracted from the API request/response example and syntactic hints such as the structure of an HTML parameters table. Additionally, we exclude the request and response examples at this stage, as they have already been utilized in generating the base OAS. Finally, all HTML attributes are removed, as they are deemed less likely to contain relevant information.

\subsection{Prompt Generation Examples}
\label{subsec:appendix-prompt-generation-examples}
\begin{figure*}
    \centering
     \includegraphics[width=1.\linewidth]{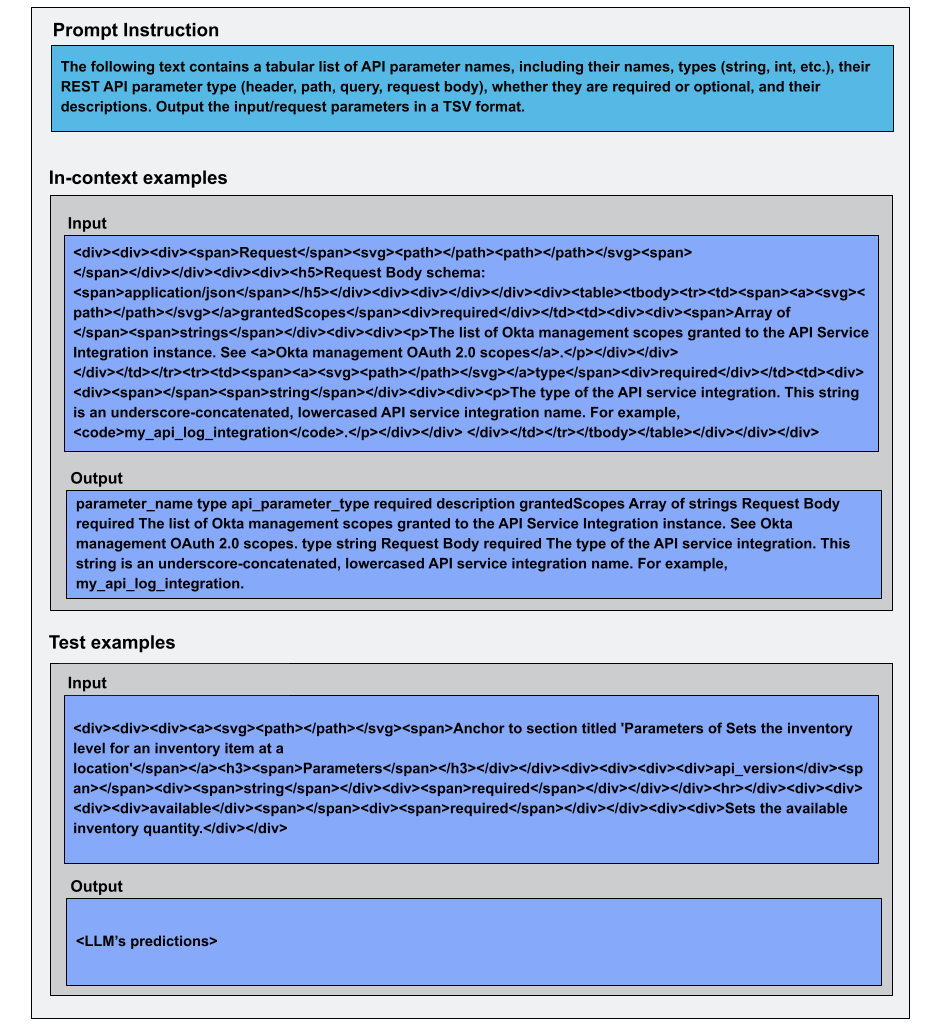}
    \caption{\label{fig:request-enrichment-prompt} An example of a prompt used for generating a TSV table for the task of request enrichment with a single in-context example.
    } 
\end{figure*}

\bibliographystyle{ACM-Reference-Format}
\bibliography{sample-base}


\begin{thebibliography}{18}

%%% the Web address.

\ifx \showCODEN    \undefined \def \showCODEN     #1{\unskip}     \fi
\ifx \showDOI      \undefined \def \showDOI       #1{#1}\fi
\ifx \showISBNx    \undefined \def \showISBNx     #1{\unskip}     \fi
\ifx \showISBNxiii \undefined \def \showISBNxiii  #1{\unskip}     \fi
\ifx \showISSN     \undefined \def \showISSN      #1{\unskip}     \fi
\ifx \showLCCN     \undefined \def \showLCCN      #1{\unskip}     \fi
\ifx \shownote     \undefined \def \shownote      #1{#1}          \fi
\ifx \showarticletitle \undefined \def \showarticletitle #1{#1}   \fi
\ifx \showURL      \undefined \def \showURL       {\relax}        \fi
% invisible to TeX
\providecommand\bibfield[2]{#2}
\providecommand\bibinfo[2]{#2}
\providecommand\natexlab[1]{#1}
\providecommand\showeprint[2][]{arXiv:#2}

\bibitem[Almazrouei et~al\mbox{.}(2023)]%
        {falcon40b}
\bibfield{author}{\bibinfo{person}{Ebtesam Almazrouei}, \bibinfo{person}{Hamza
  Alobeidli}, \bibinfo{person}{Abdulaziz Alshamsi}, \bibinfo{person}{Alessandro
  Cappelli}, \bibinfo{person}{Ruxandra Cojocaru}, \bibinfo{person}{Merouane
  Debbah}, \bibinfo{person}{Etienne Goffinet}, \bibinfo{person}{Daniel Heslow},
  \bibinfo{person}{Julien Launay}, \bibinfo{person}{Quentin Malartic},
  \bibinfo{person}{Badreddine Noune}, \bibinfo{person}{Baptiste Pannier}, {and}
  \bibinfo{person}{Guilherme Penedo}.} \bibinfo{year}{2023}\natexlab{}.
\newblock \showarticletitle{{Falcon-40B}: an open large language model with
  state-of-the-art performance}.
\newblock  (\bibinfo{year}{2023}).
\newblock


\bibitem[Andro{\v{c}}ec and Toma{\v{s}}i{\'c}(2023)]%
        {androvcec2023using}
\bibfield{author}{\bibinfo{person}{Darko Andro{\v{c}}ec} {and}
  \bibinfo{person}{Matija Toma{\v{s}}i{\'c}}.} \bibinfo{year}{2023}\natexlab{}.
\newblock \showarticletitle{Using GPT-3 to Automatically Create RESTful Service
  Descriptions}. In \bibinfo{booktitle}{\emph{2023 4th International Conference
  on Communications, Information, Electronic and Energy Systems (CIEES)}}.
  IEEE, \bibinfo{pages}{1--4}.
\newblock


\bibitem[Bahrami et~al\mbox{.}(2020)]%
        {bahrami2020deep}
\bibfield{author}{\bibinfo{person}{Mehdi Bahrami}, \bibinfo{person}{Mehdi
  Assefi}, \bibinfo{person}{Ian Thomas}, \bibinfo{person}{Wei-Peng Chen},
  \bibinfo{person}{Shridhar Choudhary}, {and} \bibinfo{person}{Hamid~R
  Arabnia}.} \bibinfo{year}{2020}\natexlab{}.
\newblock \showarticletitle{Deep sas: A deep signature-based api specification
  learning approach}. In \bibinfo{booktitle}{\emph{2020 IEEE International
  Conference on Systems, Man, and Cybernetics (SMC)}}. IEEE,
  \bibinfo{pages}{1994--2001}.
\newblock


\bibitem[Bahrami and Chen(2019)]%
        {bahrami2019watapi}
\bibfield{author}{\bibinfo{person}{Mehdi Bahrami} {and}
  \bibinfo{person}{Wei-Peng Chen}.} \bibinfo{year}{2019}\natexlab{}.
\newblock \showarticletitle{WATAPI: composing web API specification from API
  documentations through an intelligent and interactive annotation tool}. In
  \bibinfo{booktitle}{\emph{2019 IEEE International Conference on Big Data (Big
  Data)}}. IEEE, \bibinfo{pages}{4573--4578}.
\newblock


\bibitem[Bahrami and Chen(2020)]%
        {bahrami2020automated}
\bibfield{author}{\bibinfo{person}{Mehdi Bahrami} {and}
  \bibinfo{person}{Wei-Peng Chen}.} \bibinfo{year}{2020}\natexlab{}.
\newblock \showarticletitle{Automated web service specification generation
  through a transformation-based learning}. In
  \bibinfo{booktitle}{\emph{Services Computing--SCC 2020: 17th International
  Conference, Held as Part of the Services Conference Federation, SCF 2020,
  Honolulu, HI, USA, September 18--20, 2020, Proceedings 17}}. Springer,
  \bibinfo{pages}{103--119}.
\newblock


\bibitem[Cao et~al\mbox{.}(2017)]%
        {cao2017automated}
\bibfield{author}{\bibinfo{person}{Hanyang Cao}, \bibinfo{person}{Jean-R{\'e}my
  Falleri}, {and} \bibinfo{person}{Xavier Blanc}.}
  \bibinfo{year}{2017}\natexlab{}.
\newblock \showarticletitle{Automated generation of REST API specification from
  plain HTML documentation}. In \bibinfo{booktitle}{\emph{Service-Oriented
  Computing: 15th International Conference, ICSOC 2017, Malaga, Spain, November
  13--16, 2017, Proceedings}}. Springer, \bibinfo{pages}{453--461}.
\newblock


\bibitem[Danielsen and Jeffrey(2013)]%
        {danielsen2013}
\bibfield{author}{\bibinfo{person}{Peter~J. Danielsen} {and}
  \bibinfo{person}{Alan Jeffrey}.} \bibinfo{year}{2013}\natexlab{}.
\newblock \showarticletitle{Validation and Interactivity of Web API
  Documentation}. In \bibinfo{booktitle}{\emph{2013 IEEE 20th International
  Conference on Web Services}}. \bibinfo{pages}{523--530}.
\newblock
\urldef\tempurl%
\url{https://doi.org/10.1109/ICWS.2013.76}
\showDOI{\tempurl}


\bibitem[Huang et~al\mbox{.}(2024)]%
        {huang2024generating}
\bibfield{author}{\bibinfo{person}{Ruikai Huang}, \bibinfo{person}{Manish
  Motwani}, \bibinfo{person}{Idel Martinez}, {and} \bibinfo{person}{Alessandro
  Orso}.} \bibinfo{year}{2024}\natexlab{}.
\newblock \showarticletitle{Generating REST API Specifications through Static
  Analysis}.
\newblock  (\bibinfo{year}{2024}).
\newblock


\bibitem[Jiang et~al\mbox{.}(2023)]%
        {jiang2023mistral}
\bibfield{author}{\bibinfo{person}{Albert~Q. Jiang}, \bibinfo{person}{Alexandre
  Sablayrolles}, \bibinfo{person}{Arthur Mensch}, \bibinfo{person}{Chris
  Bamford}, \bibinfo{person}{Devendra~Singh Chaplot}, \bibinfo{person}{Diego
  de~las Casas}, \bibinfo{person}{Florian Bressand}, \bibinfo{person}{Gianna
  Lengyel}, \bibinfo{person}{Guillaume Lample}, \bibinfo{person}{Lucile
  Saulnier}, \bibinfo{person}{Lélio~Renard Lavaud},
  \bibinfo{person}{Marie-Anne Lachaux}, \bibinfo{person}{Pierre Stock},
  \bibinfo{person}{Teven~Le Scao}, \bibinfo{person}{Thibaut Lavril},
  \bibinfo{person}{Thomas Wang}, \bibinfo{person}{Timothée Lacroix}, {and}
  \bibinfo{person}{William~El Sayed}.} \bibinfo{year}{2023}\natexlab{}.
\newblock \bibinfo{title}{Mistral 7B}.
\newblock
\newblock
\showeprint[arxiv]{2310.06825}~[cs.CL]


\bibitem[Jiang et~al\mbox{.}(2024)]%
        {jiang2024mixtral}
\bibfield{author}{\bibinfo{person}{Albert~Q. Jiang}, \bibinfo{person}{Alexandre
  Sablayrolles}, \bibinfo{person}{Antoine Roux}, \bibinfo{person}{Arthur
  Mensch}, \bibinfo{person}{Blanche Savary}, \bibinfo{person}{Chris Bamford},
  \bibinfo{person}{Devendra~Singh Chaplot}, \bibinfo{person}{Diego de~las
  Casas}, \bibinfo{person}{Emma~Bou Hanna}, \bibinfo{person}{Florian Bressand},
  \bibinfo{person}{Gianna Lengyel}, \bibinfo{person}{Guillaume Bour},
  \bibinfo{person}{Guillaume Lample}, \bibinfo{person}{Lélio~Renard Lavaud},
  \bibinfo{person}{Lucile Saulnier}, \bibinfo{person}{Marie-Anne Lachaux},
  \bibinfo{person}{Pierre Stock}, \bibinfo{person}{Sandeep Subramanian},
  \bibinfo{person}{Sophia Yang}, \bibinfo{person}{Szymon Antoniak},
  \bibinfo{person}{Teven~Le Scao}, \bibinfo{person}{Théophile Gervet},
  \bibinfo{person}{Thibaut Lavril}, \bibinfo{person}{Thomas Wang},
  \bibinfo{person}{Timothée Lacroix}, {and} \bibinfo{person}{William~El
  Sayed}.} \bibinfo{year}{2024}\natexlab{}.
\newblock \bibinfo{title}{Mixtral of Experts}.
\newblock
\newblock
\showeprint[arxiv]{2401.04088}~[cs.LG]


\bibitem[Kim et~al\mbox{.}(2022)]%
        {kim2022}
\bibfield{author}{\bibinfo{person}{Myeongsoo Kim}, \bibinfo{person}{Qi Xin},
  \bibinfo{person}{Saurabh Sinha}, {and} \bibinfo{person}{Alessandro Orso}.}
  \bibinfo{year}{2022}\natexlab{}.
\newblock \showarticletitle{Automated test generation for REST APIs: no time to
  rest yet}. In \bibinfo{booktitle}{\emph{Proceedings of the 31st ACM SIGSOFT
  International Symposium on Software Testing and Analysis}} (, Virtual, South
  Korea,) \emph{(\bibinfo{series}{ISSTA 2022})}.
  \bibinfo{publisher}{Association for Computing Machinery},
  \bibinfo{address}{New York, NY, USA}, \bibinfo{pages}{289–301}.
\newblock
\showISBNx{9781450393799}
\urldef\tempurl%
\url{https://doi.org/10.1145/3533767.3534401}
\showDOI{\tempurl}


\bibitem[Martin-Lopez et~al\mbox{.}(2021)]%
        {martin2021black}
\bibfield{author}{\bibinfo{person}{Alberto Martin-Lopez},
  \bibinfo{person}{Andrea Arcuri}, \bibinfo{person}{Sergio Segura}, {and}
  \bibinfo{person}{Antonio Ruiz-Cort{\'e}s}.} \bibinfo{year}{2021}\natexlab{}.
\newblock \showarticletitle{Black-box and white-box test case generation for
  RESTful APIs: Enemies or allies?}. In \bibinfo{booktitle}{\emph{2021 IEEE
  32nd International Symposium on Software Reliability Engineering (ISSRE)}}.
  IEEE, \bibinfo{pages}{231--241}.
\newblock


\bibitem[Martin-Lopez et~al\mbox{.}(2022)]%
        {martin2022online}
\bibfield{author}{\bibinfo{person}{Alberto Martin-Lopez},
  \bibinfo{person}{Sergio Segura}, {and} \bibinfo{person}{Antonio
  Ruiz-Cort{\'e}s}.} \bibinfo{year}{2022}\natexlab{}.
\newblock \showarticletitle{Online testing of RESTful APIs: Promises and
  challenges}. In \bibinfo{booktitle}{\emph{Proceedings of the 30th ACM Joint
  European Software Engineering Conference and Symposium on the Foundations of
  Software Engineering}}. \bibinfo{pages}{408--420}.
\newblock


\bibitem[Rozière et~al\mbox{.}(2024)]%
        {rozière2024code}
\bibfield{author}{\bibinfo{person}{Baptiste Rozière}, \bibinfo{person}{Jonas
  Gehring}, \bibinfo{person}{Fabian Gloeckle}, \bibinfo{person}{Sten Sootla},
  \bibinfo{person}{Itai Gat}, \bibinfo{person}{Xiaoqing~Ellen Tan},
  \bibinfo{person}{Yossi Adi}, \bibinfo{person}{Jingyu Liu},
  \bibinfo{person}{Romain Sauvestre}, \bibinfo{person}{Tal Remez},
  \bibinfo{person}{Jérémy Rapin}, \bibinfo{person}{Artyom Kozhevnikov},
  \bibinfo{person}{Ivan Evtimov}, \bibinfo{person}{Joanna Bitton},
  \bibinfo{person}{Manish Bhatt}, \bibinfo{person}{Cristian~Canton Ferrer},
  \bibinfo{person}{Aaron Grattafiori}, \bibinfo{person}{Wenhan Xiong},
  \bibinfo{person}{Alexandre Défossez}, \bibinfo{person}{Jade Copet},
  \bibinfo{person}{Faisal Azhar}, \bibinfo{person}{Hugo Touvron},
  \bibinfo{person}{Louis Martin}, \bibinfo{person}{Nicolas Usunier},
  \bibinfo{person}{Thomas Scialom}, {and} \bibinfo{person}{Gabriel Synnaeve}.}
  \bibinfo{year}{2024}\natexlab{}.
\newblock \bibinfo{title}{Code Llama: Open Foundation Models for Code}.
\newblock
\newblock
\showeprint[arxiv]{2308.12950}~[cs.CL]


\bibitem[Sohan et~al\mbox{.}(2015)]%
        {sohan2015spyrest}
\bibfield{author}{\bibinfo{person}{Sheikh~Mohammed Sohan},
  \bibinfo{person}{Craig Anslow}, {and} \bibinfo{person}{Frank Maurer}.}
  \bibinfo{year}{2015}\natexlab{}.
\newblock \showarticletitle{Spyrest: Automated restful API documentation using
  an HTTP proxy server (N)}. In \bibinfo{booktitle}{\emph{2015 30th IEEE/ACM
  International Conference on Automated Software Engineering (ASE)}}. IEEE,
  \bibinfo{pages}{271--276}.
\newblock


\bibitem[Touvron et~al\mbox{.}(2023)]%
        {touvron2023llama}
\bibfield{author}{\bibinfo{person}{Hugo Touvron}, \bibinfo{person}{Louis
  Martin}, \bibinfo{person}{Kevin Stone}, \bibinfo{person}{Peter Albert},
  \bibinfo{person}{Amjad Almahairi}, \bibinfo{person}{Yasmine Babaei},
  \bibinfo{person}{Nikolay Bashlykov}, \bibinfo{person}{Soumya Batra},
  \bibinfo{person}{Prajjwal Bhargava}, \bibinfo{person}{Shruti Bhosale},
  \bibinfo{person}{Dan Bikel}, \bibinfo{person}{Lukas Blecher},
  \bibinfo{person}{Cristian~Canton Ferrer}, \bibinfo{person}{Moya Chen},
  \bibinfo{person}{Guillem Cucurull}, \bibinfo{person}{David Esiobu},
  \bibinfo{person}{Jude Fernandes}, \bibinfo{person}{Jeremy Fu},
  \bibinfo{person}{Wenyin Fu}, \bibinfo{person}{Brian Fuller},
  \bibinfo{person}{Cynthia Gao}, \bibinfo{person}{Vedanuj Goswami},
  \bibinfo{person}{Naman Goyal}, \bibinfo{person}{Anthony Hartshorn},
  \bibinfo{person}{Saghar Hosseini}, \bibinfo{person}{Rui Hou},
  \bibinfo{person}{Hakan Inan}, \bibinfo{person}{Marcin Kardas},
  \bibinfo{person}{Viktor Kerkez}, \bibinfo{person}{Madian Khabsa},
  \bibinfo{person}{Isabel Kloumann}, \bibinfo{person}{Artem Korenev},
  \bibinfo{person}{Punit~Singh Koura}, \bibinfo{person}{Marie-Anne Lachaux},
  \bibinfo{person}{Thibaut Lavril}, \bibinfo{person}{Jenya Lee},
  \bibinfo{person}{Diana Liskovich}, \bibinfo{person}{Yinghai Lu},
  \bibinfo{person}{Yuning Mao}, \bibinfo{person}{Xavier Martinet},
  \bibinfo{person}{Todor Mihaylov}, \bibinfo{person}{Pushkar Mishra},
  \bibinfo{person}{Igor Molybog}, \bibinfo{person}{Yixin Nie},
  \bibinfo{person}{Andrew Poulton}, \bibinfo{person}{Jeremy Reizenstein},
  \bibinfo{person}{Rashi Rungta}, \bibinfo{person}{Kalyan Saladi},
  \bibinfo{person}{Alan Schelten}, \bibinfo{person}{Ruan Silva},
  \bibinfo{person}{Eric~Michael Smith}, \bibinfo{person}{Ranjan Subramanian},
  \bibinfo{person}{Xiaoqing~Ellen Tan}, \bibinfo{person}{Binh Tang},
  \bibinfo{person}{Ross Taylor}, \bibinfo{person}{Adina Williams},
  \bibinfo{person}{Jian~Xiang Kuan}, \bibinfo{person}{Puxin Xu},
  \bibinfo{person}{Zheng Yan}, \bibinfo{person}{Iliyan Zarov},
  \bibinfo{person}{Yuchen Zhang}, \bibinfo{person}{Angela Fan},
  \bibinfo{person}{Melanie Kambadur}, \bibinfo{person}{Sharan Narang},
  \bibinfo{person}{Aurelien Rodriguez}, \bibinfo{person}{Robert Stojnic},
  \bibinfo{person}{Sergey Edunov}, {and} \bibinfo{person}{Thomas Scialom}.}
  \bibinfo{year}{2023}\natexlab{}.
\newblock \bibinfo{title}{Llama 2: Open Foundation and Fine-Tuned Chat Models}.
\newblock
\newblock
\showeprint[arxiv]{2307.09288}~[cs.CL]


\bibitem[Yang et~al\mbox{.}(2018a)]%
        {yang2018automatically}
\bibfield{author}{\bibinfo{person}{Jinqiu Yang}, \bibinfo{person}{Erik
  Wittern}, \bibinfo{person}{Annie~TT Ying}, \bibinfo{person}{Julian Dolby},
  {and} \bibinfo{person}{Lin Tan}.} \bibinfo{year}{2018}\natexlab{a}.
\newblock \showarticletitle{Automatically extracting web api specifications
  from html documentation}.
\newblock \bibinfo{journal}{\emph{arXiv preprint arXiv:1801.08928}}
  (\bibinfo{year}{2018}).
\newblock


\bibitem[Yang et~al\mbox{.}(2018b)]%
        {Yang-2018}
\bibfield{author}{\bibinfo{person}{Jinqiu Yang}, \bibinfo{person}{Erik
  Wittern}, \bibinfo{person}{Annie T.~T. Ying}, \bibinfo{person}{Julian Dolby},
  {and} \bibinfo{person}{Lin Tan}.} \bibinfo{year}{2018}\natexlab{b}.
\newblock \showarticletitle{Towards Extracting Web API Specifications from
  Documentation}. In \bibinfo{booktitle}{\emph{Proceedings of the 15th
  International Conference on Mining Software Repositories}} (Gothenburg,
  Sweden) \emph{(\bibinfo{series}{MSR '18})}. \bibinfo{publisher}{Association
  for Computing Machinery}, \bibinfo{address}{New York, NY, USA},
  \bibinfo{pages}{454–464}.
\newblock
\showISBNx{9781450357166}
\urldef\tempurl%
\url{https://doi.org/10.1145/3196398.3196411}
\showDOI{\tempurl}


\end{thebibliography}








\end{document}